# News Sentiment Analysis

Antony Samuels, John Mcgonical

University of Southern California, Caltech

aantonysamuels@gmail.com  to.john.mcgonical@gmail.com

*Abstract* — **Modern technological era has reshaped traditional lifestyle in several domains. The medium of publishing news and events has become faster with the advancement of Information Technology (IT). IT has also been flooded with immense amounts of data, which is being published every minute of every day, by millions of users, in the shape of comments, blogs, news sharing through blogs, social media micro-blogging websites and many more. Manual traversal of such huge data is a challenging job; thus, sophisticated methods are acquired to perform this task automatically and efficiently. News reports events that comprise of emotions – good, bad, neutral. Sentiment analysis is utilized to investigate human emotions (i.e., sentiments) present in textual information. This paper presents a lexicon-based approach for sentiment analysis of news articles. The experiments have been performed on BBC news dataset, which expresses the applicability and validation of the adopted approach.**

*Keywords— Sentiment analysis, Lexicon-based approach, news articles.*

## I. INTRODUCTION

With the emergence of the Internet, web and mobile technologies, people have changed their way of consuming news. Traditional physical newspapers and magazines have been replaced by virtual online versions like online news and weblogs. Readers are more inclined to use online sources of news mainly due to two key features: interactivity and immediacy [1].

In this day and age, people want to consume as much news, from as many sources, as they possibly can, on matters that are important to them or matters that catch their attention. Interactivity refers to the inherent tendency depicted by the masses that makes them consume news of their interest. Immediacy is a feature that represents the need of people to be informed about news with no delay in time [2]. The world we live in and the technology we are accustomed to, allows people to benefit from these features by providing them instant news on events as they happen in real-time. Online news websites have developed effective strategies to draw peoples' attention [3]. Online news expresses opinions regarding news entities, which may comprise of people, places or even things, while reporting on events that have recently occurred [4]. For this reason interactive emotion rating services are offered by various channels of several news websites, i.e., news can be positive, negative or neutral [5].

Sentiment Analysis or Opinion Mining is a way of finding out the polarity or strength of the opinion (positive or negative) that is expressed in written text, in the case of this paper – a news article [3] [4]. Manual labeling of sentiment words is a time consuming process. There are two popular approaches that are utilized to automate the process of sentiment analysis. The first process makes use of a lexicon of weighted words and the second process is based on approaches of machine learning. Lexicon based methods use a word stock dictionary with opinion words and match given set of words in a text for finding polarity. As opposed to machine learning methods, this approach does not need to preprocess data not does it have to train a classifier [6]. This research is based on a method for Lexicon-based sentiment analysis of news articles

The remainder of this paper is organized as follows: Section II presents related work conducted in sentiment analysis for news articles. Section III presents the proposed methodology and experiment setup of this paper. Results have been presented in Section IV followed by limitations of the research in Section V. Finally, Section VI presents the conclusion of this research.

## II. RELATED WORK

Many researchers have contributed in news sentiment analysis using different approaches. A brief discussion on the work done previously on sentiment analysis is provided in this section.

Reis, Olmo Benevenuto, Prates and An proposed a methodology to discover the relationship between sentiment polarity and news popularity [3]. Using different sentiment analysis methods, an experiment was conducted by utilizing the content of 69,907 headlines generated by four most reputed media corporations –The New York Times, BBC, Reuters, and Dailymail. Extracting features from text of news headlines, the research analyzed the sentiment polarity of these headlines. The research concluded that the polarity of the headline had a great impact on the popularity of the news article. The research found that negative and positive news headlines gained greater interest than news headlines that had a neutral tone.

Godbole, Srinivasaiah, and Sekine built an algorithm based on sentiment lexicons which could help in finding the sentiment words and entities associated in the text corpus of news and blogs by looking at the co-occurrence of entity and sentiment word in the same sentence [4]. Seven dimensions comprising of general, health, crime, sports, business, politics and media were selected for sentiment analysis from news and blogs. Two trends were analyzed in the experiment - 1) Polarity: sentiment associated with entity is positive or negative and 2) Subjectivity: how much sentiment an entity garners. Score for both polarity and subjectivity were calculated.

Islam, Ashraf, Abir and Mottalib proposed an approach to classify online news. Sentiment analysis was done at sentence level and a dynamic dictionary with predefined positive and negative words was used in order to get help for finding sentiment polarity [6]. Following steps was carried out for news article classification. 1) Selection of an online news article. 2) Extraction of sentences from the news articles. Sentences can be simple, compound, complex and compound complex. 3) Search for positive words, phrases or clauses in those sentences and finding their polarities. 4) Combining the polarities of all sentences to get the final polarity of news article. 91% accurate results were collected for classification of news articles.

Meyer, Bikdash, Dai performed fine grained sentiment analysis of financial news headline using machine learning approach and lexicon based approach. A total of eight experiments were conducted to find more accurate results. Results from both approaches were also compared [8]. For lexicon based approach, Bag of Words (BOW) model along with General Inquirer Lexicon (H4N) was used to determine sentiment polarities. For machine learning approach, Parts of Speech (POS) syntactic model was used. In the experiments concluded that more accurate results were obtained by using machine learning approach.

Shirsat, Jagdale and Deshmukh proposed a methodology for sentiment analysis on document level so that polarity of an entire news article could be determined [9]. The paper explored a dataset of 2225 documents. After text pre-processing through tokenization, stop word removal and stemming, post-processing was done on the entire news article. In this step the sentiment score of the article was assigned based on this sentiment score. The news articles were categorized as positive, negative or neutral.

Agarwal, Sharma, Sikka and Dhir performed opinion mining using python packages to classify words and SentiWordNet 3.0 to identify the positive and negative words so that total impact i.e. positive or negative sentiment in news headline can be evaluated [10]. The impact of news headlines has been analyzed using two algorithms.

Algorithm 1: Preprocessing of each word

Select news headline, then pre-process each word in it using POS tagger and perform Lemmatization, and Stemming. This is done using Natural Language Tool Kit (NLTK).

Algorithm 2: Analyzing news headlines

After pre-processing pass each word in to SentiWordNet 3.0 dictionary to find positive, negative and objective scores. If positive score > negative score then mark news headline as positive. And if positive score < negative score then mark news headline as negative.

Lei, Rao, Li, Quan, and Wenyin have built a model for detecting social emotions induced by news articles, tweets etc. [11]. The model comprises of modules for document selection, tagging of parts of speech, and lexicon generation based on social emotions. This model first creates a training set from corpus of news documents then applies techniques of POS tagging and feature extraction. After this step social emotion lexicons have been generated through calculation of the probabilities of the emotions based on the document. To test the accuracy of the model, a dataset of 40,897 news articles collected from the societal channel has been used.

III. RESEARCH METHODOLOGY

The methodology used for sentiment analysis of news articles in this paper is based on the Lexicon-based approach. Sentiment analysis can generally be carried out using supervised or unsupervised approaches. A supervised approach comprises of a set of labeled training data that is used to build a classification model with the intent of using this model to classify new data for which labels are not present. Unsupervised or Lexicon-based approaches to sentiment analysis do not require any training data. In this approach, the sentiments conveyed by a word are inferred on grounds of the polarity of the word. In case of a sentence or a document, the polarities of the individual words that compose the document collectively convey the sentiment of the sentence or the document. Thus the polarity of a sentence is the accumulative total (sum) of polarities of the individual words (or phrases) in the sentence [12].

This approach utilizes some predefined lists of words such that each word in the list is associated with a specific sentiment. Further this approach can use the following methods:

1. *Dictionary-based methods:* in these methods lexicon dictionary is used in order to find out the positive opinion words and negative opinion words.

2. *Corpus-based methods:* in these methods large corpus of words is used and based on syntactic patterns other opinion words can be found within the context.

Sentiment analysis can be done on document level, sentence level, word level or phrase level. This paper explores sentiment analysis on the document level. Similar to [13] [14], this research identifies whether the documents new articles expressed opinions are positive, negative or neutral. The dictionary based approach has been used for sentiment analysis of news articles utilizing the wordNet lexical dictionary. The experiment for this research was carried out using the Rapid Miner tool. The methodology for this experiment has been presented in Fig.1.

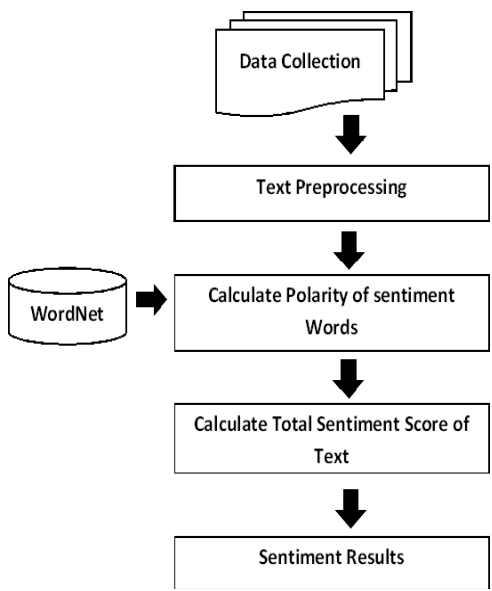

Fig. 1. Sentiment Analysis Methodology

The methodology comprised of 5 steps, starting with data collection. The BBC News dataset has been used for this experiment. The next step was preprocessing the collected data in order to reduce inconsistencies in the dataset. The polarity of the words in the collected news articles was computed next using the wordNet lexical dictionary. The steps have been explained in detail below.

*A. Data Collection*

The BBC News dataset was utilized for this experiment. The dataset is available online at http://mlg.ucd.ie/datasets/bbc.html. This data set comprises of a total of 2225 documents that comprise of news articles reported on the BBC news website between the years 2004-2005. The news stories belong to 5 (five) topical areas. The dataset comprises of the following class labels: business, entertainment, politics, sport, and tech.

*B. Text Pre-processing*

News articles in the dataset were preprocessed. Preprocessing is a necessary step to clean text (lessen noise of text) and to reduce inconsistencies from it so that this cleansed data can more effectively be utilized in text mining or sentiment analysis task [15]. The entire preprocessing task was carried out using the Rapid miner tool which provides a vast set of operators for preprocessing tasks. The first preprocessing task was tokenizing the text in news articles into a set of tokens by using the "Tokenize" operator. Tokenizing breaks a sequence of sentences (combination of strings) into individual components such as words, phrases or symbols which are termed tokens. Apart from individual words and phrases, tokens can even comprise of entire sentences. During tokenization some characters, such as punctuation marks, are discarded. After tokenization, the text of the entire documents was changed to a lower case format using the "Transform cases" operator. Stop words from the text were removed using "filter stop word (English)" operator. The next task was reducing inflected or derived words through a process called Stemming. Stemming of words was done using the "stem (wordNet)" operator.

*C. Calculate Polarity of Sentiment of Sentiment words*

After preprocessing, the statistical technique known as Term Frequency-Inverse Document Frequency (TF-IDF) has been used. In TF-IDF term frequency is counted [16]. According to this technique words that occur frequently in a document are considered important and a weight is given to these words. Using TF-IDF important words or terms in a document were identified and assigned a weightage according to the occurrence of various words in the news article.

After identification of important words, a dictionary has been used for assigning sentiment score to the discovered words. The WordNet dictionary, which is also known as a lexical database for English language, has been used in this experiment. WordNet contains more than 118,000 different word forms and more than 90,000 different word senses [17]. WordNet provides accurate results to find opinion words in a given text and to give sentiment score to them.

*D. Calculate Total Sentiment Score*

According to the principle of document level sentiment analysis, each individual document is tagged with its respective polarity. This is generally done by finding polarities of each individual words/phrases and sentences and combining them to predict the polarity of whole document. Treating each new article as a document, the sentiment conveyed in the article has been computed by combining polarities of individual words/phrases and sentences in news articles.

The sentiment score of whole news article has been calculated using the "extract sentiment" operator. This operator provides final results about sentiments: text having a sentiment score of -1 is considered negative and text having a sentiment score of +1 is positive. This operator provided accurate results by using SentiWordNet 3.0.0 dictionary which is actually an extension of the wordNet dictionary. WordNet and SentiWordNet are connected by Synset IDs. Also by using Score sentiment function based on WordNet and SentiWordNet dictionary, total sentiment score of news article was calculated.

*E. Sentiment Results*

News articles were classified in to positive, negative and neutral classes by looking at their total sentiment score. News articles sentiment was then calculated as the average value of total word sentiments.

## IV. RESULTS AND DISCUSSION

News articles having a sentiment score of 0 were considered as neutral and news articles with a score of +1 were treated as positive whereas news articles having a sentiment score of -1 have been treated as negative. The results of the experiment have been presented in Table 1.

TABLE I. SENTIMENT RESULTS

| NEWS CLASS | TOTAL ARTICLES | POSITIVE | NEGATIVE | NEUTRAL |
|---|---|---|---|---|
| Business | 510 | 274 | 205 | 31 |
| Entertainment | 401 | 163 | 220 | 18 |
| Politics | 417 | 205 | 200 | 12 |
| Sport | 511 | 246 | 236 | 29 |
| Tech | 401 | 170 | 216 | 15 |

It was observed that a majority of new articles fell into the negative or positive categories with a minor percentage of articles having neutral sentiments. A majority of news articles in the Entertainment and Tech category exhibited negative sentiments, whereas the categories of business and sports comprised of a majority of articles depicting positive sentiments. The category of politics had almost an equal proportion of articles exhibiting positive as well as negative sentiments. The results of sentiment analysis have been graphically represented in Fig.2.

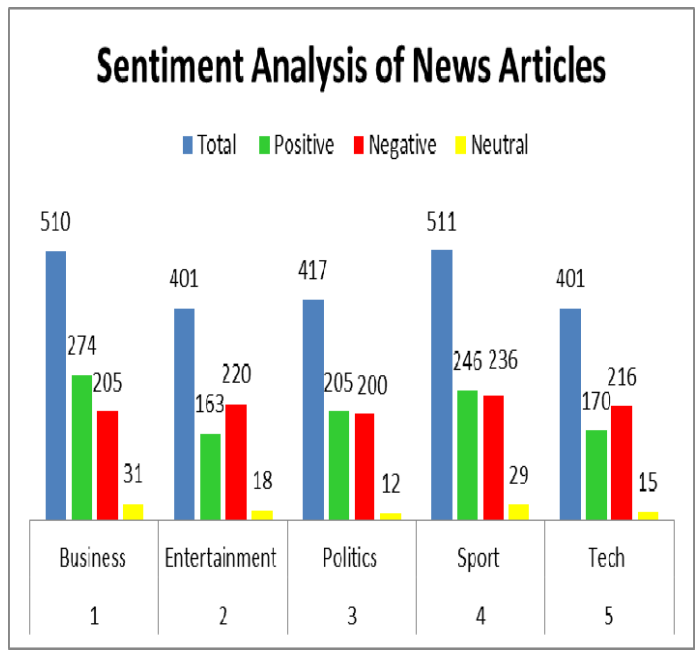

Fig. 2. Results of New Articles

V. RESEARCH LIMITATIONS AND CHALLENGES

Sentiment analysis focuses on text written in English and Chinese with few researches now being carried out on Arabic, Thai and Italian. Insufficient or limited word coverage as many new words and their new semantics must be updated in lexical database [18]. The accuracy of sentiment classification is also a challenging task in sentiment analysis. Finalizing the techniques most suitable for specific sentiment analysis tasks is also a challenge as the nature of the dataset keeps changing - datasets of news, reviews, and blogs all have text expressed in various formats. This causes a variation in the accuracy and performance of sentiment analysis classifiers [19].

For the proposed model a limitation is that it only uses English news articles from one source for sentiment analysis.

VI. CONCLUSION

There are many directions in sentiment analysis that can be explored. This paper explored sentiment analysis of news and blogs using a dataset from BBC comprising of new articles between the year 2004 and 2005. It was observed that categories of business and sports had more positive articles, whereas entertainment and tech had a majority of negative articles. Future work in this regard will be based on sentiment analysis of news using various machine learning approaches with the development of an online application from where users can read news of their interests. Also, based on sentiment analysis methods, readers can customize their news feed.